\documentclass[10pt,twocolumn,letterpaper]{article}

\usepackage{iccv}
\usepackage{times}
\usepackage{epsfig}
\usepackage{graphicx}
\usepackage{amsmath}
\usepackage{amssymb}
\usepackage{booktabs}
\usepackage{multirow}
\usepackage{dcolumn}
\usepackage{makecell}

\usepackage[breaklinks=true,bookmarks=false]{hyperref}

\iccvfinalcopy 

\def\iccvPaperID{6004} 

\ificcvfinal\fi

\begin{document}

\title{Fast Convergence of DETR with Spatially Modulated Co-Attention}

\author{Peng Gao$^{1}$ \quad Minghang Zheng$^{3}$ \quad Xiaogang Wang$ ^{2}$ \quad Jifeng Dai$^{4}$ \quad Hongsheng Li$^{2}$\\
$^1$Shanghai AI Laboratory, \\ $^2$CUHK-SenseTime Joint Laboratory, The Chinese University of Hong Kong \\ $^3$Peking University \quad $^4$SenseTime Research\\
{\tt\small 1155102382@link.cuhk.edu.hk} \quad
{\tt\small hsli@ee.cuhk.edu.hk}

}

\maketitle
\ificcvfinal\thispagestyle{empty}\fi

\begin{abstract}
   The recently proposed Detection Transformer (DETR) model successfully applies Transformer to objects detection and achieves comparable performance with two-stage object detection frameworks, such as Faster-RCNN. However, DETR suffers from its slow convergence.
   Training DETR~\cite{carion2020end} from scratch needs 500 epochs to achieve a high accuracy. To accelerate its convergence, we propose a simple yet effective scheme for improving the DETR framework, namely Spatially Modulated Co-Attention (SMCA) mechanism. The core idea of SMCA is to conduct location-aware co-attention in DETR by constraining co-attention responses to be high near initially estimated bounding box locations. Our proposed SMCA increases DETR's convergence speed by replacing the original co-attention mechanism in the decoder while keeping other operations in DETR unchanged. Furthermore, by integrating multi-head and scale-selection attention designs into SMCA, our fully-fledged SMCA can achieve better performance compared to DETR with a dilated convolution-based backbone (45.6 mAP at 108 epochs vs. 43.3 mAP at 500 epochs). We perform extensive ablation studies on COCO dataset to validate SMCA. Code is released at \url{https://github.com/gaopengcuhk/SMCA-DETR}.
\end{abstract}

\section{Introduction}

The recently proposed DETR~\cite{carion2020end} has significantly simplified object detection pipeline by removing hand-crafted anchor~\cite{ren2016faster} and non-maximum suppression (NMS)~\cite{bodla2017soft}. However, the convergence speed of DETR is slow compared with two-stage~\cite{girshick2015region,girshick2015fast,ren2016faster} or one-stage~\cite{liu2016ssd,redmon2016you,lin2017focal} detectors (500 vs. 40 epochs). Slow convergence of DETR makes it difficult for researchers to further extend the algorithm and thus hinders its widespread usage. 

\begin{figure}
    \centering
    \scalebox{0.9}[0.8]{
    \includegraphics[width=\linewidth]{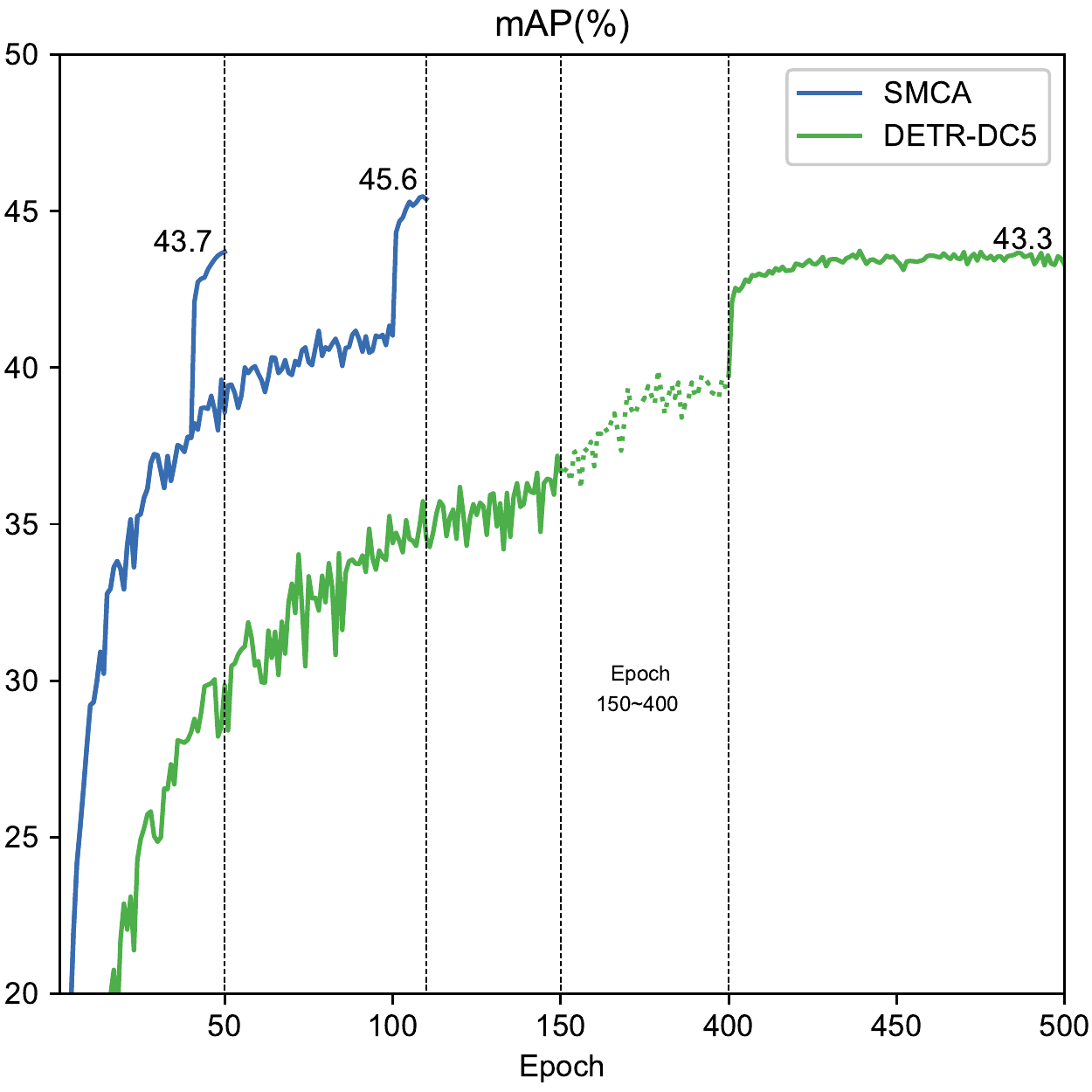}
    }
    \caption{Comparison of DETR-DC5 trained for 500 epochs, and our proposed SMCA trained for 50 epochs and 108 epochs. The convergence speed of the proposed SMCA is faster than DETR.}
    \label{figure:mAP}
\end{figure}

In DETR, there are a series of object query vectors responsible for detecting objects at different spatial locations. Each object query interacts with the spatial visual features encoded by a Convolution Neural Network (CNN)~\cite{he2016deep}, adaptively collects information from spatial locations with a co-attention mechanism, and then estimates the bounding box locations and object categories. However, in the decoder of DETR, the co-attended visual regions for each object query might be unrelated to the bounding box to be predicted by the query. Thus the decoder of DETR needs long training epochs to search for the properly co-attended regions to accurately identify the corresponding objects.

Motivated by this observation, we propose a novel module named Spatially Modulated Co-attention (SMCA), which is a plug-and-play module to replace the existing co-attention mechanism in DETR and achieves faster convergence and improved performance with simple modifications. The proposed SMCA dynamically predicts initial center and scale of the box corresponding to each object query to generate a 2D spatial Gaussian-like weight map. The weight map is element-wisely multiplied with the co-attention feature maps of object query and image features to more effectively aggregate query-related information from the visual feature map. In this way, the spatial weight map effectively modulates the search range of each object query's co-attention to be properly around the initially estimated object center and scale. By leveraging the predicted Gaussian-distributed spatial prior, our SMCA can significantly speed up the training of DETR.

Although naively incorporating the spatially-modulated co-attention mechanism into DETR speeds up the convergence, the performance is worse compared with DETR (41.0 mAP at 50 epochs, 42.7 at 108 epochs vs. 43.3 mAP at 500 epochs). Motivated by the effectiveness of multi-head attention-based Transformer~\cite{vaswani2017attention} and multi-scale feature~\cite{lin2017feature} in previous research work, our SMCA is further augmented with the multi-scale visual feature encoding in the encoder and the multi-head attention in the decoder.
For multi-scale visual feature encoding in the encoder, instead of naively rescaling and upsampling the multi-scale features from the CNN backbone to form a joint multi-scale feature map, Intra-scale and multi-scale self-attention mechanisms are introduced to directly and efficiently propagate information between the visual features of multiple scales. For the proposed multi-scale self-attention, visual features at all spatial locations of all scales interact with each other via self-attention. However, as the number of all spatial locations at all scales is quite large and leads to large computational cost, we introduce the intra-scale self-attention to alleviate the heavy computation. The properly combined intra-scale and multi-scale self-attention achieve efficient and discriminative multi-scale feature encoding. In the decoder, each object query can adaptively select features of proper scales via the proposed scale-selection attention. For the multiple co-attention heads in the decoder, all heads estimate head-specific object centers and scales to generate a series of different spatial weight maps for spatially modulating the co-attention features. Each of the multiple heads aggregates visual information from slightly different locations and thus improves the detection performance. 

Our SMCA is motivated by the following research. DRAW~\cite{gregor2015draw} proposed a differential read-and-write operator with dynamically predicted Gaussian sampling points for image generation. Deformable DETR~\cite{zhu2020deformable} achieved fast convergence of DETR with learnable sparse sampling. Compared with Deformable DETR, our proposed SMCA explores another direction for fast convergence of DETR by exploring dynamic Gaussian-like spatial prior. Besides, SMCA can accelerate the training of DETR by only replacing co-attention in the decoder. Deformable DETR replaces the Transformer with deformable attention for encoder and decoder. SMCA demonstrates that exploring global information can also result in the fast convergence of DETR. 

We summarize our contributions below: 1) We propose a novel Spatial Modulated Co-Attention (SMCA), which can accelerate the convergence of DETR by conducting location-constrained object regression. SMCA is a plug-and-play module in the original DETR. The basic version of SMCA without multi-scale features and multi-head attention can already achieve 41.0 mAP at 50 epochs and 42.7 mAP at 108 epochs. It takes 265 V100 GPU hours to train the basic version of SMCA for 50 epochs. 2) Our full SMCA further integrates multi-scale features and multi-head spatial modulation, which can further significantly improve and surpass DETR with much fewer training iterations. SMCA can achieve 43.7 mAP at 50 epochs and 45.6 mAP at 108 epochs, while DETR-DC5 achieves 43.3 mAP at 500 epochs. It takes 600 V100 GPU hours to train the full SMCA for 50 epochs. 3) We perform extensive ablation studies on COCO 2017 dataset to validate the proposed SMCA module and the network design.

\section{Related Work}
\noindent \textbf{Object Detection.}
Motivated by the success of deep learning on image classification~\cite{krizhevsky2017imagenet,he2016deep}, deep learning has been successfully applied to object detection~\cite{girshick2015region}. Deep learning-based object detection frameworks can be categorized into two-stage, one-stage, and end-to-end ones. For two-stage object detectors including RCNN~\cite{girshick2015region}, Fast RCNN~\cite{girshick2015fast} and Faster RCNN~\cite{ren2016faster}, the region proposal layer generates a few regions from dense sliding windows first, and the ROI align~\cite{he2017mask} layer then extracts fine-grained features and perform classification over the pooled features. For one-stage detectors such as YOLO~\cite{redmon2016you} and SSD~\cite{liu2016ssd}, they conduct object classification and location estimation directly over dense sling windows. Both two-stage and one-stage methods need complicated post-processing to generate the final bounding box predictions. 

Recently, another branch of object detection methods~\cite{stewart2016end, salvador2017recurrent,ren2017end, carion2020end} beyond one-stage and two-stage ones has gained popularity. They directly supervise bounding box predictions end-to-end with Hungarian bipartite matching. However, DETR~\cite{carion2020end} suffered from slow convergence compared with two-stage and one-stage object detectors.
Deformable DETR~\cite{zhu2020deformable} accelerates the convergence speed of DETR via learnable sparse sampling coupled with multi-scale deformable encoder. TSP~\cite{sun2020rethinking} combined RCNN- or FCOS-based methods with DETR. Deformable DETR, TSP-RCNN and TSP-FCOS only explored local information while our SMCA explores global information with a self-attention mechanism. UP-DETR~\cite{dai2020up} propose a novel self-supervised loss to enhance the convergence speed.


\noindent \textbf{Transformer.}
CNN~\cite{lecun1998gradient} and LSTM~\cite{hochreiter1997long} can be used for modeling sequential data. CNN processes input sequences with a weight-shared sliding window manner. LSTM processes inputs with a recurrence mechanism controlled by several dynamically predicted gating functions. Transformer~\cite{vaswani2017attention} introduces a new architecture beyond CNN and LSTM by performing information exchange between all pairs of input using key-query value attention. Transformer has achieved success on machine translation, after which Transformer has been adopted in different fields, including model pre-training~\cite{devlin2018bert,radford2018improving,radford2019language,brown2020language}, visual recognition~\cite{parmar2019stand,dosovitskiy2020image}, and multi-modality fusion~\cite{yu2019deep,gao2019dynamic,lu2019vilbert,gao2019multi,geng2021dynamic}. Transformer has quadratic complexity for information exchange between all pairs of inputs, which is difficult to scale up for longer input sequences. Many methods have been proposed to tackle this problem. Reformer~\cite{kitaev2020reformer} proposed a reversible FFN and clustering self-attention. Linformer~\cite{wang2020linformer} and FastTransformer~\cite{katharopoulos2020transformers} proposed to remove the softmax in the transformer and perform matrix multiplication between query and value first to obtain a linear-complexity transformer. LongFormer~\cite{beltagy2020longformer} perform self-attention within a local window instead of the whole input sequence. Container~\cite{gao2021container} is a recently proposed backbone network which unify convolution and self-attention mechanism though contextual aggregation. Transformer has been utilized in DETR to enhance the features by performing feature exchange between different positions and object query. In SMCA, intra-scale multi-scale self-attention is utilized for multi-scale information exchange.



\section{Spatially Modulated Co-Attention}

\subsection{Overview}

\begin{figure*}
    \centering
    \scalebox{0.9}[0.78]{
    \includegraphics[width=\linewidth]{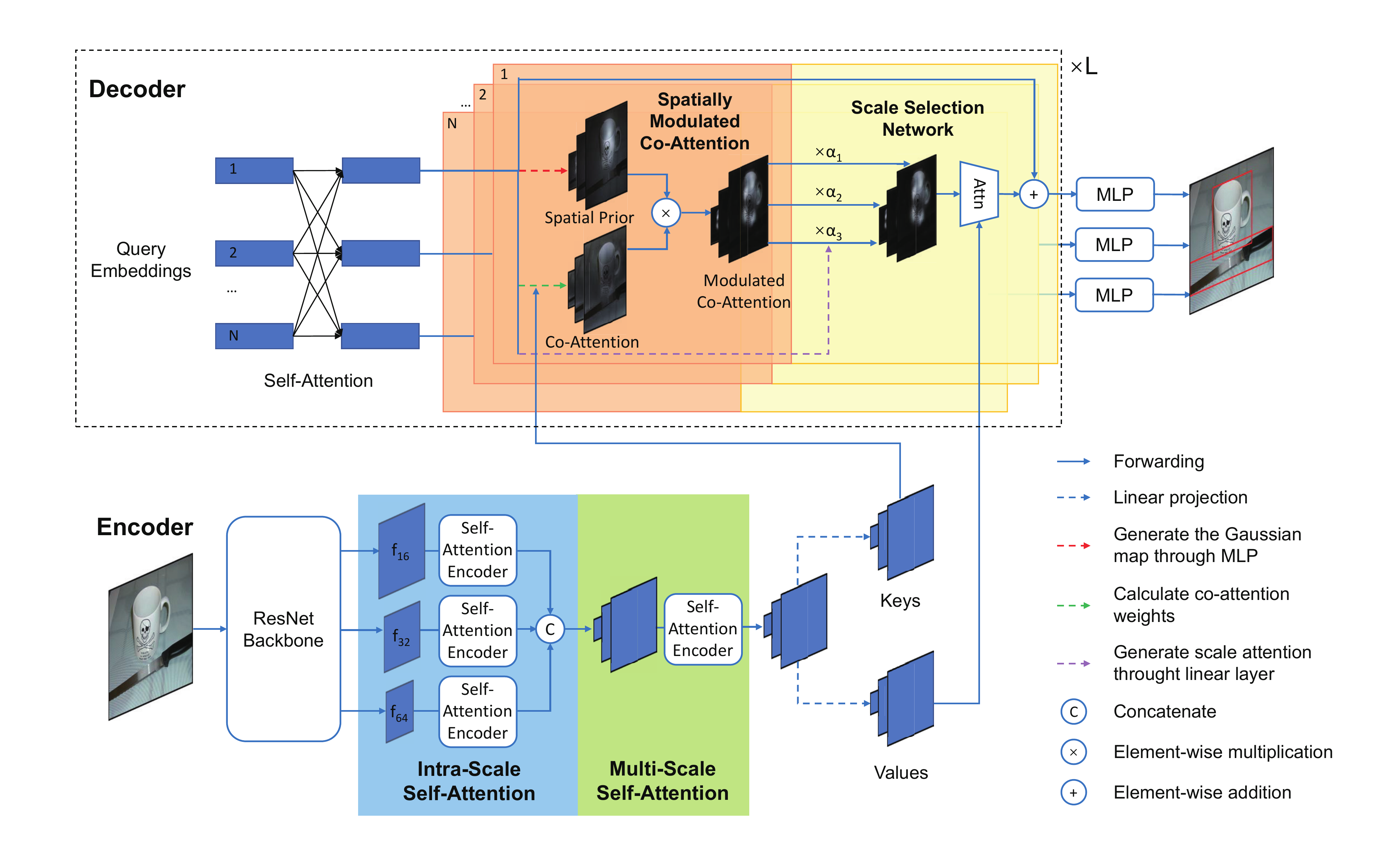}
    }
    \caption{The overall pipeline of Spatially Modulated Co-Attention (SMCA) with intra-scale self-attention, multi-scale self-attention, spatial modulation, and scale-selection attention modules. Each object query performs spatially modulated co-attention and then predicts the target bounding boxes and their object categories. $N$ stands for the number of object queries. $L$ stands for the layers of decoder.}
    \label{figure:gmca}
\end{figure*}

In this section, we will first revisit the basic design of DETR~\cite{carion2020end} and then introduce the basic version of SMCA. After introducing SMCA, we will introduce how to integrate multi-head and scale-selection attention mechanisms into SMCA. SMCA is illustrated in Figure ~\ref{figure:gmca}.

\subsection{A Revisit of DETR}

End-to-end object DEtection with TRansformer (DETR)~\cite{carion2020end} formulates detection as a set prediction problem. Convolution Neural Network (CNN)~\cite{he2016deep} extracts visual feature maps $f  \in \mathbb{R}^{C \times H \times W}$ from an image $I \in \mathbb{R}^{3 \times H_0 \times W_0}$, where $H,W$ and $H_0,W_0$ are the height/width of the image and the feature, respectively.

The visual features augmented with position embedding $f_{pe}$ would be fed into the encoder. Self-attention would be applied to $f_{pe}$ to generate the key, query, and value features $K, Q, V$ to exchange information between features at all spatial positions. To increase the feature diversity, such features would be split into multiple groups along the channel dimension for the multi-head self-attention. The multi-head normalized dot-product attention is conducted as
\begin{align}
    E_{i} &= \operatorname{Softmax}(K_{i}^T Q_{i} / {\sqrt{d}})V_{i}, \label{eq:self-attention} \\
    E &= \operatorname{Concat}(E_1, \dots, E_{H}), \nonumber
\end{align}
where $K_i, Q_i, V_i$ denote the $i$th feature group of the key, query, and value features. There are $H$ groups for each type of features, and the output encoder features $E$ is then further transformed and input into the decoder of the Transformer.

Given the visual feature $E$ encoded from the encoder, DETR performs co-attention between object queries $O_q \in \mathbb{R}^{N \times C}$ and the visual features $E \in \mathbb{R}^{L\times C}$, where $N$ denotes the number of pre-specified object queries and $L$ is the number of the spatial visual features.
\begin{align}
    Q &= \operatorname{FC}(O_q), \,\, K, V = \operatorname{FC}(E) \nonumber \\
    C_{i} &= \operatorname{Softmax}(K_{i}^T Q_{i} / \sqrt{d})V_{i}, \label{eq:co-attention} \\
    C &= \operatorname{Concat}(C_1, \dots, C_{H}), \nonumber
\end{align}
where $\operatorname{FC}$ denotes a single-layer linear transformation, and $C_i$ denotes the co-attended feature for the object query $O_q$ from the $i$th co-attention head. The decoder's output features of each object query is then further transformed by a Multi-Layer Perceptron (MLP) to output class score and box location for each object. 

Given box and class prediction, the Hungarian algorithm is applied between predictions and ground-truth box annotations to identify the learning targets of each object query.

\subsection{Spatially Modulated Co-Attention}

The co-attention in DETR is unaware of the predicted bounding boxes and thus requires many iterations to learn how to generate the proper attention map for each object query. The core idea of our SMCA is to combine the learnable co-attention maps with handcrafted query spatial priors, which constrain the attended features to be around the object queries' initial estimations and thus to be more related to the final predictions. The SMCA module is illustrated in the Figure~\ref{figure:gmca} in orange.

\vspace{2pt}
\noindent {\bf Dynamic spatial weight maps.} Each object query first dynamically predicts the center and scale of its responsible object, which are then used to generate a 2D Gaussian-like spatial weight map. The center of the Gaussian-like distribution are parameterized in the normalized coordinates of $[0,1] \times [0,1]$.
The initial prediction of the normalized center $c^\mathrm{norm}_h, c^\mathrm{norm}_w$ and scale $s_h, s_w$ of the Gaussian-like distribution for object query $O_q$ is formulated as
\begin{align}
     c^\mathrm{norm}_h, c^\mathrm{norm}_w &= \operatorname{sigmoid}(\operatorname{MLP}(O_q)), \label{eq:scale}\\
     s_h, s_w &= \operatorname{FC}(O_q) \nonumber,
\end{align}
where the object query $O_q$ is projected to obtain normalized prediction center in the two dimensions $c^{\mathrm{norm}}_h, c^{\mathrm{norm}}_w$ with a 2-layer MLP followed by a sigmoid activation function. The predicted center is then unnormalized to obtain the center coordinates $c_h, c_w$ in the original image.
$O_q$ would also dynamically estimate the object scales $s_h, s_w$ along the two dimensions to create the 2D Gaussian-like weight map, which is then used to re-weight the co-attention map to emphasize features around the predicted object location.

Objects in natural images show diverse scales and height/width ratios. The design of predicting width- and height-independent $s_h, s_w$ can better tackle the complex object aspect ratios in real-world. For large or small objects, SMCA dynamically generates $s_h, s_w$ of different values, so that the modulated co-attention map by the spatial weight map $G$ can aggregate sufficient information from all parts of large objects or suppress background clutters for the small objects.
After predicting the object center $c_w, c_h$ and scale $s_w, s_h$, SMCA generates the Gaussian-like weight map as
\begin{align}
    G(i,j) = \operatorname{exp}\left(-\frac{(i-c_w)^2}{\beta s_w^2} -\frac{(j-c_h)^2}{\beta s_h^2}\right),
\end{align}
where $(i,j) \in [0,W] \times [0,H]$ is the spatial indices of the weight map $G$, and $\beta$ is a hyper-parameter to modulate the bandwidth of the Gaussian-like distribution.
The weight map $G$ generally assigns high importance to locations near the center and low importance to positions far from the center.
$\beta$ can be manually tuned with a handcrafted scheme to ensure $G$ covering enough spatial range at the beginning so that the network can receive more informative gradients.

\vspace{2pt}
\noindent {\bf Spatially-modulated co-attention.} Given the dynamically generated spatial prior $G$, we modulate the co-attention maps $C_i$ between object query $O_q$ and self-attention encoded feature $E$ with the spatial prior $G$. For each co-attention map $C_i$ generated with the dot-product attention (Eq. (\ref{eq:co-attention})), we modulate the co-attention maps $C_i$ with the spatial weight map $G$, where $G$ is shared for all co-attention heads in the basic version of our SMCA,
\begin{align}
    C_{i} = \operatorname{softmax}(K_{i}^T Q_{i} / \sqrt{d} + \log G) V_{i}. \label{eq:single_head_gmca}
\end{align}
Our SMCA performs element-wise addition between the logarithm of the spatial map $G$ and the dot-product co-attention $K_{i}^T Q_{i} / \sqrt{d}$ followed by softmax normalization over all spatial locations. By doing so, the decoder co-attention would weight more around the predicted bounding box locations, which limits the search space of the spatial patterns of the co-attention and thus increases the convergence speed. The Gaussian-like weight map is illustrated in Figure~\ref{figure:gmca}, which constrains the co-attention to focus more on regions near the predicted bounding box location and thus significantly increases the convergence speed. 

\vspace{2pt}
\noindent {\bf SMCA with multi-head modulation.} We also investigate to modulate co-attention features differently for different co-attention heads. Each head starts from a head-shared center $[c_w, c_h]$, similar to that of the basic version of SMCA, and then predicts a head-specific center offset $[\Delta c_{w,i}, \Delta c_{h, i}]$ and head-specific scales $s_{w,i}, s_{h,i}$. The Gaussian-like spatial weight map $G_i$ can thus be generated based on the head-specific center $[c_w +\Delta c_{w,i}, c_h + \Delta c_{h,i}]$ and scales $s_{w,i}, s_{h,i}$. The co-attention feature maps $C_1, \dots, C_H$ of $H$ heads can be obtained as
\begin{align}
    C_{i} = \operatorname{softmax}(K_{i}^T Q_{i} / \sqrt{d} + \log G_i) V_{i}. \label{eq:multi_head_gmca}
\end{align}
Different from Eq. (\ref{eq:single_head_gmca}) that shares $\operatorname{log}G$ for all attention heads, the above Eq. (\ref{eq:multi_head_gmca}) modulates co-attention maps by head-specific spatial weight maps $\operatorname{log}G_i$. The multiple spatial weight maps can emphasize diverse context and improve the detection accuracy.

\vspace{2pt}
\noindent{\bf SMCA with multi-scale visual features.} Feature pyramid is popular in object detection frameworks and generally leads to significant improvements over single-scale feature encoding. Motivated by the FPN~\cite{lin2017feature} in previous works, we also integrate multi-scale features into SMCA. The basic version of SMCA conducts co-attention between object queries and single-scale feature maps. As objects naturally have different scales, we can further improve the framework by replacing single-scale feature encoding with multi-scale feature encoding in the encoder of the Transformer.

Given an image, the CNN extracts the multi-scale visual features with downsampling rates 16, 32, 64 to obtain multi-scale features $f_{16}$, $f_{32}$, $f_{64}$, respectively. The multi-scale features are directly obtained from the CNN backbone. For multi-scale self-attention encoding in the encoder, features at all locations of different scales are treated equally. The self-attention mechanism propagates and aggregates information between all feature pixels of different scales. However, the number of feature pixels of all scales is quite large and the multi-scale self-attention operation is therefore computationally costly. To tackle the issue, we introduce the intra-scale self-attention encoding as an auxiliary operator to assist the multi-scale self-attention encoding. Specifically, dot-product attention is used to propagate and aggregate features only between feature pixels within each scale. The weights of the Transformer block (with self-attention and feedforward sub-networks) are shared across different scales. Our empirical study shows that parameter sharing across scales enhances the generalization capability of intra-scale self-attention encoding. For the final design of the encoder in SMCA, it adopts 2 blocks of intra-scale encoding, followed by 1 block of multi-scale encoding, and another 2 blocks of intra-scale encoding. The design has a very similar detection performance to that of 5 blocks of multi-scale self-attention encoding with less computation. 

Given the encoded multi-scale features $E_{16}$, $E_{32}$, $E_{64}$ with downsampling rates of 16, 32, 64, a naive solution for the decoder to perform co-attention would be first re-scaling and concatenating the multi-scale features to form a single-scale feature map, and then conducting co-attention between object query and the resulting feature map.
However, we notice that some queries might only require information from a specific scale but not always from all the scales. 
For example, the information for small objects is missing in low-resolution feature map $E_{64}$. Thus the object queries responsible for small objects should more effectively acquire information only from high-resolution feature maps. On the other hand, traditional methods, such as FPN, assigns each bounding box explicitly to the feature map of a specific scale.
Different from FPN~\cite{lin2017feature}, we propose to automatically select scales for each box using learnable scale-attention attention.
Each object query generates scale-selection attention weights as
\begin{align}
    \alpha_{16}, \alpha_{32}, \alpha_{64} = \operatorname{Softmax}(\operatorname{FC}(O_q)),
    \label{eq:scale_selection}
\end{align}
\textcolor{black}{where $\alpha_{16}$, $\alpha_{32}$, $\alpha_{64}$ stand for the importance of selecting $f_{16}$, $f_{32}$, $f_{64}$. To conduct co-attention between the object query $O_q$ and the multi-scale visual features $E_{16}, E_{32}, E_{64}$, we first obtain the multi-scale key and value features $K_{i,16}, K_{i,32}, K_{i,64}$ and $V_{i,16}, V_{i,32}, V_{i,64}$ for attention head $i$, respectively, from $E_{16}$, $E_{32}$, $E_{64}$ with separate linear projections. 
To conduct co-attention for each head $i$ between $O_q$ and key/value features of each scale $j \in \{16, 32, 64\}$, the spatially-modulated co-attention in Eq. (\ref{eq:single_head_gmca}) is adaptively weighted and aggregated by the scale-selection weights $\alpha_{16}, \alpha_{32}, \alpha_{64}$ as}
\begin{align}
    C_{i,j} &= \operatorname{Softmax}(K_{i,j}^T Q_{i} / \sqrt{d} + \operatorname{log}{G_i}) V_{i,j}  \odot \alpha_j, \\
    C_{i} &= \sum_{\text{all } j} C_{i,j}, \quad \text{ for } \, j \in \{16, 32, 64\},
\end{align}
\textcolor{black}{where $C_{i,j}$ stands for the co-attention features between the $i$th co-attention head between query and visual features of scale $j$.
$C_{i,j}$'s are weightedly aggregated according to the scaled attention weights $\alpha_j$ obtained in Eq. (\ref{eq:scale_selection}).
With such a scale-selection attention mechanism, the scale most related to each object query is softly selected.}




\begin{table}\footnotesize
\small
\setlength\tabcolsep{2pt}
    \centering
    \scalebox{0.9}[0.8]{
    \begin{tabular}{c|cccccccc}
        \toprule
        Method & Epochs & time(s) & GFLOPs & mAP & AP$_S$ & AP$_M$ & AP$_L$\\
        \midrule
        DETR &500 & 0.038 &86 & 42.0 &20.5& 45.8&61.1 &\\
        DETR-DC5 &500 & 0.079 &187 & 43.3 & 22.5&47.3 &61.1\\
        \midrule
        \makecell[c]{SMCA w/o\\ multi-scale} & 50 &  0.043 & 86 & 41.0 & 21.9& 44.3&59.1\\
        \makecell[c]{SMCA w/o\\ multi-scale}& 108 &  0.043 & 86 & 42.7 & 22.8& 46.1&60.0\\
        SMCA & 50 &  0.100 & 152& 43.7 &24.2 &47.0 &60.4\\
        SMCA & 108 &  0.100 & 152& 45.6 &25.9 &49.3 &62.6\\
        \bottomrule
    \end{tabular}
    }
    \vspace{2pt}
    \caption{Comparison with DETR model over training epochs, mAP, inference time and GFLOPs.}
    \label{tab:tab3}
\end{table}

\section{Experiments}

\subsection{Experiment setup}

\noindent \textbf{Dataset.} We validate our proposed SMCA over COCO 2017~\cite{lin2014microsoft} dataset. Specifically, we train on COCO 2017 training dataset and validate on the validation dataset, which contains 118k and 5k images, respectively. We report mAP for performance evaluation following previous research~\cite{carion2020end}.

\vspace{2pt}
\noindent \textbf{Implementation details.} We follow the experiment setup in the original DETR~\cite{carion2020end}. We denote the features extracted by ResNet-50~\cite{he2016deep} as SMCA-R50. Different from DETR, we use 300 object queries instead of 100 and replace the original cross-entropy classification loss with focal loss~\cite{lin2017focal}. The initial probability of focal loss is set as 0.01 to stabilize the training process.

We report the performance trained for 50 epochs and the learning rate decreases to 1/10 of its original value at the 40th epoch. The learning rate is set as $10^{-4}$ for the Transformer encoder-encoder and $10^{-5}$ for the pre-trained ResNet backbone and optimized by AdamW optimizer~\cite{loshchilov2018fixing}. 

For multi-scale feature encoding, we use downsampling ratios of 16, 32, 64 by default. \textcolor{black}{For bipartite matching~\cite{stewart2016end,carion2020end}, the coefficients of classification loss, L1 distance loss, GIoU loss is set as 2, 5, 2, respectively. After bounding box assignment via bipartite matching, SMCA is trained by minimizing the classification loss, bounding box L1 loss, and GIoU loss with coefficients 2, 5, 2, respectively.} For Transformer layers~\cite{vaswani2017attention}, we use post-norm similar to those in previous approaches \cite{carion2020end}. We use random crop for data augmentation with the largest width or height set as 1333 for all experiments following~\cite{carion2020end}. All models are trained on 8 V100 GPUs with 1 image per GPU.

\subsection{Comparison with DETR}
 SMCA shares the same architecture with DETR except for the proposed new co-attention modulation in the decoder and an extra linear network for generating the spatial modulation prior. The increase of computational cost of SMCA and training time of each epoch are marginal. For SMCA with single-scale features (denoted as ``SMCA w/o multi-scale''), we keep the dimension of self-attention to be 256 and the intermediate dimension of FFN to be 2048. For SMCA with multi-scale features, we set the intermediate dimension of FFN to be 1024 and use 5 layers of intra-scale and multi-scale self-attention in the encoder to have similar amount of parameters and fair comparison with DETR. As shown in Table ~\ref{tab:tab3}, the performance of ``SMCA w/o multi-scale'' reaches 41.0 mAP with single-scale features and 43.7 mAP with multi-scale features at 50 epochs. Given longer training procedure, mAP of SMCA increases from 41.0 to 42.7 with single-scale features and from 43.7 to 45.6 with multi-scale features. ``SMCA w/o multi-scale" can achieve better AP$_s$ and AP$_M$ compared with DETR. SMCA can achieve better overall performance on objects of all scales by considering multi-scale information and the proposed spatial modulation. The convergence speed of SMCA is 10 times faster than DETR-based methods.

Given the significant increase of convergence speed and performance, the FLOPs and the increase of inference time of SMCA are marginal. With single-scale features, the inference time increases from $0.038s\rightarrow0.041s$ and FLOPs increase by 0.06G. With multi-scale features, the inference speed increase from $0.079s\rightarrow0.100s$, while the GFLOPs actually decrease because our multi-scale SMCA only uses 5 layers of self-attention layers for the encoder. Thin layers in the Transformer and convolution without dilation in the last stage of ResNet backbone achieve similar efficiency as the original dilated DETR model.

\subsection{Ablation Study}

To validate different components of our proposed SMCA, we perform ablation studies on the importance of the proposed spatial modulation, multi-head vs. head-shared modulation, and multi-scale encoding and scale-selection attention in comparison with the baseline DETR.

\begin{table}\footnotesize
    \centering
    \small
    \setlength\tabcolsep{5pt}
    \scalebox{0.9}[0.8]{
    \begin{tabular}{cc|c|c|c}
        \toprule
         \multicolumn{2}{c|}{Method}&mAP&AP50&AP75  \\
         \midrule
         Baseline & DETR-R50 & 34.8 & 56.2 & 36.9\\
         \midrule
         \multirow{3}*{\makecell[c]{Head-shared Spatial \\Modulation}} &+Indep. (bs8) & 40.2& 61.4&42.7 \\
         &+Indep. (bs16) & 40.2& 61.3&42.9 \\
         &+Indep. (bs32) & 39.9& 61.0&42.4 \\
         \midrule
         \multirow{3}*{\makecell[c]{Multi-head Spatial \\ Modulation}} 
         & +Fixed&38.5 & 60.7&40.2 \\
         & +Single& 40.4&61.8 &43.3 \\
         & +Indep. & 41.0&62.2 &43.6 \\
         \bottomrule
    \end{tabular}
    }
    \vspace{2pt}
    \caption{Ablation study on the importance of spatial modulation, multi-head mechanism. mAP, AP50, and AP75 are reported on COCO 2017 validation set.}
    \label{tab:tab1}
\end{table}

\begin{table}\footnotesize
    \centering
    \small
    \setlength\tabcolsep{5pt}
    \scalebox{0.9}[0.78]{
    \begin{tabular}{cc|c|c}
        \toprule
         \multicolumn{2}{c|}{Method} & mAP & Params (M) \\
         \midrule
         \multicolumn{2}{c|}{\makecell[c]{SMCA}} & 41.0 & 41.0 \\
         \midrule
         \multicolumn{2}{c|}{\makecell[c]{SMCA\\(2Intra-Multi-2Intra)}} & 43.7 & 39.5 \\
         \midrule
         \multicolumn{2}{c|}{\makecell[c]{SMCA w/o SSA \\(2Intra-Multi-2Intra)}}  &42.6&39.5 \\
         \midrule
         \multicolumn{2}{c|}{3Intra} &42.9&37.9 \\
         \multicolumn{2}{c|}{3Multi} &43.3&37.9 \\
         \multicolumn{2}{c|}{5Intra} &43.3&39.5 \\
         \midrule
         \multirow{3}*{Weight Share} & Shared FFN &43.0&42.2\\
         & Shared SA&42.8&44.7 \\
         & No Share&42.3&47.3\\
         \bottomrule
    \end{tabular}
    }
    \vspace{2pt}
    \caption{Ablation study on the importance of combining intra-scale and multi-scale propagation, and the weight sharing for intra-scale self-attention. ``Shared FFN'' stands for only sharing weights of the feedfoward network of intra-scale self-attention. ``Shared SA'' stands for sharing the weights of the self-attention network. ``No share'' stands for no weight sharing in intra-scale self attention.}
    \label{tab:tab2}
\end{table}

\begin{table*}\footnotesize
    \centering
    \small
    \scalebox{0.88}[0.78]{
    \begin{tabular}{c|cccc|cccccc}
        \toprule
         Model&Epochs&GFLOPs&Params (M)&AP&AP$_{50}$&AP$_{75}$&AP$_S$&AP$_M$&AP$_L$  \\
         \midrule
         DETR-R50~\cite{carion2020end}&500 &86& 41& 42.0&62.4 & 44.2& 20.5& 45.8& 61.1 \\
         DETR-DC5-R50~\cite{carion2020end}&500&187& 41&43.3 & 63.1 & 45.9 & 22.5 & 47.3 & 61.1 \\
         Faster RCNN-FPN-R50~\cite{carion2020end}&36& 180&  42&40.2 & 61.0 & 43.8& 24.2& 43.5& 52.0 \\
         Faster RCNN-FPN-R50++~\cite{carion2020end}&108&180 & 42&42.0 &62.1 &45.5 &26.6 &45.4 &53.4 \\
         Deformable DETR-R50 (Single-scale)~\cite{zhu2020deformable} &50 & 78&  34&39.7 &60.1 &42.4 &21.2 &44.3 &56.0 \\
         Deformable DETR-R50 (50 epochs)~\cite{zhu2020deformable} &50&173 & 40 &43.8 &62.6 &47.7 &26.4 &47.1 &58.0 \\
         Deformable DETR-R50 (150 epochs)~\cite{zhu2020deformable} &150&173& 40 &45.3 &64.3 &49.1 &27.1 &48.4 &60.0 \\
         UP-DETR-R50 ~\cite{dai2020up}& 150 &86& 41&40.5 &60.8 &42.6 & 19.0& 44.4 & 60.0\\
         UP-DETR-R50+~\cite{dai2020up} & 300 &86& 41&42.8 &63.0 &45.3 & 20.8& 47.1 & 61.7\\
         TSP-FCOS-R50 ~\cite{sun2020rethinking}& 36 &189& 52&43.1 &62.3 &47.0 & 26.6& 46.8 & 55.9\\
         TSP-RCNN-R50~\cite{sun2020rethinking} & 36 &188& 64&43.8 &63.3 &48.3 & 28.6& 46.9 & 55.7\\
         TSP-RCNN+-R50~\cite{sun2020rethinking} & 96 &188& 64&45.0 &64.5 &\textbf{49.6} & \textbf{29.7}& 47.7 & 58.0\\
         SMCA-Container(Single-scale)~\cite{gao2021container} & 50 & 86 & 38 & 44.2 & \textbf{66.1} & 47.3 & 23.8 & 47.9 & \textbf{63.1} \\
         \midrule
         SMCA-R50&50& 152& 40 &43.7 &63.6&47.2&24.2 &47.0 &60.4 \\
         SMCA-R50&108& 152& 40&\textbf{45.6} & 65.5& 49.1&25.9 &\textbf{49.3} & 62.6 \\
         \midrule
         DETR-R101~\cite{carion2020end}&500&152 & 60 &43.5&63.8 & 46.4& 21.9& 48.0& 61.8 \\
         DETR-DC5-R101~\cite{carion2020end}&500&253 & 60 &44.9 & 64.7 & 47.7 & 23.7 & 49.5 & \textbf{62.3} \\
         Faster RCNN-FPN-R101~\cite{carion2020end}&36&256& 60 &42.0 &62.1 &45.5 &26.6 &45.4 &53.4 \\
         Faster RCNN-FPN-R101+~\cite{carion2020end}&108&246& 60 &44.0& 63.9& 47.8& 27.2&48.1 &56.0 \\
         TSP-FCOS-R101~\cite{sun2020rethinking} & 36 &255& 70 &44.4 &63.8 &48.2 & 27.7& 48.6 & 57.3\\
         TSP-RCNN-R101~\cite{sun2020rethinking} & 36 &254& 83 &44.8 &63.8 &49.2 & 29.0& 47.9 & 57.1\\
         TSP-RCNN+-R101~\cite{sun2020rethinking} & 96 &254& 83 &\textbf{46.5} &66.0 &\textbf{51.2} & \textbf{29.9}& 49.7 & 59.2\\
         \midrule
         SMCA-R101 & 50 & 218 & 58 & 44.4 & 65.2 & 48.0 & 24.3 & 48.5 & 61.0 \\
         SMCA-R101 & 108 & 218 & 58 & 46.3 & \textbf{66.6} & 50.2 & 27.2 & \textbf{50.5} & \textbf{63.2} \\
         \bottomrule
    \end{tabular}
    }
    \caption{Comparison with DETR-like object detectors on COCO 2017 validation set.}
    \label{tab:final}
\end{table*}

\vspace{2pt}
\noindent \textbf{The baseline DETR model.}
We choose DETR with ResNet-50 backbone as our baseline model. It is trained for 50 epochs with the learning rate dropping to 1/10 of the original value at the 40th epoch. Different from the original DETR, we increase the object query from 100 to 300 and replace the original cross entropy loss with the focal loss. As shown in Table~\ref{tab:tab1}, the baseline DETR model can achieve an mAP of 34.8 at 50 epochs.

\vspace{2pt}
\noindent \textbf{Head-shared spatially modulated co-attention.}
Based on the baseline DETR, we first test adding a head-shared spatial modulation as specified in Eq.~(\ref{eq:single_head_gmca}) by keeping factors including the learning rate, training schedule, self-attention parameters, and coefficients of the loss to be the same as the baseline. The spatial weight map is generated based on the predicted height and width shared for all heads contain height- and width-independent scale prediction to better tackle the scale variance problem. We denote the method by ``Head-shared Spatial Modulation + Indep.'' in Table~\ref{tab:tab1}. The performance increases from 34.8 to 40.2 compared with baseline DETR. The large performance gain (+5.4) validates the effectiveness of SMCA, which not only accelerates the convergence speed of DETR but also improve its performance by a large margin. We further test the performance of head-shared spatial modulation with different batch sizes of 8, 16, and 32 as shown in Table~\ref{tab:tab1}. The results show that our SMCA is insensitive to different batch sizes.

\vspace{2pt}
\noindent \textbf{Multi-head vs. head-shared spatially modulated co-attention.}
For spatial modulation with multiple heads of separate predictable scales, all heads in Transformer are modulated by different spatial weight maps $G_i$ following Eq.~(\ref{eq:multi_head_gmca}). All heads start from the same object center and predict offsets w.r.t. the common center and head-specific scales. The design of multi-head spatial modulation for co-attention enables the model to learn diverse attention patterns simultaneously. After switching from head-shared spatial modulation to multi-head spatial modulation (denoted by ``Multi-head Spatial Modulation + Indep.'' in Table~\ref{tab:tab1}), the performance increases from 40.2 to 41.0 compared with the head-shared modulated co-attention in SMCA. The importance of multi-head mechanism has also been discussed in Transformer~\cite{vaswani2017attention}.

\vspace{2pt}
\noindent \textbf{Design of multi-head spatial modulation for co-attention.}
We test whether the width and height scales of the spatial weight maps should be manually set, shared, or independently predicted. 
As shown in Table~\ref{tab:tab1}, we test fixed-scale Gaussian-like spatial map (only predicting the center and fixing the scale of the Gaussian-like distribution to be the constant 1). The fixed-scale spatial modulation results in a 38.5 mAP (denoted as ``+Fixed''), which has +3.7 gain over the baseline DETR-R50 and validates the effectiveness of predicting centers for spatial modulation to constrain the co-attention. As objects in natural images have varying sizes, scales can be predicted to adapt to objects of different sizes. Thus we allow the scale to be a single predictable variable as in Eq.~(\ref{eq:scale}). If 
such a single predictable scale for spatial modulation (denoted by ``+Single''), SMCA can achieve 40.4 mAP and is +1.9 compared with the above fixed-scale modulation. By further predicting independent scales for height and width, our SMCA can achieve 41.0 mAP (denoted by ``+Indep.''), which is +0.6 higher compared with the SMCA with a single predictable scale. The results demonstrate the importance of predicting both height and width scales for the proposed spatial modulation. We visualize the co-attention patterns in the supplementary materials, which show that independent spatial modulation can generate more accurate and compact co-attention patterns compared with fixed-scale and shared-scale spatial modulation.

\vspace{2pt}
\noindent \textbf{Multi-scale feature encoding and scale-selection attention.}
The above SMCA only conducts co-attention between single-scale feature maps and the object query. As objects in natural images exist in different scales, we conduct multi-scale feature encoding in the encoder via adopting 2 layers of intra-scale self-attention, followed by 1 layer of multi-scale self-attention, and then another 2 layers of intra-scale self-attention. We denote the above design by ``SMCA (2Intra-Multi-2Intra)''. As shown in Table~\ref{tab:tab2}, we start from SMCA with a single-scale visual feature map, which achieves 41.0 mAP. After integrating multi-scale features with the 2intra-multi-2intra self-attention design, the performance can be enhanced from 41.0 to 43.7. As we introduce 3 convolutions to project features output from ResNet-50 to 256 dimensions, we make the hidden dimension of FFN decrease from 2048 to 1024 and the number of encoder layer decrease from 6 to 5 to make the parameter comparable to other models. \textcolor{black}{To validate the effectiveness of scale-selection attention (SSA), we perform ablation studies on SMCA without integrating SSA (denoted by ``SMCA w/o SSA''). As shown in Table~\ref{tab:tab2}, SMCA w/o SSA decreases the performance from 43.7 to 42.6.}

After validating the effectiveness of the proposed multi-scale feature encoding and scale-selection attention module, we further validate the effectiveness of the design of 2intra-multi-2intra-scale self-attention. By switching the 2intra-multi-2intra design to simply stacking 5 intra-scale self-attention layers, the performance drops from 43.7 to 43.3, due to the lack of cross-scale information exchange. 5 layers of intra-scale self-attention (denoted by ``5Intra'') encoder achieves better performance than 3Intra self-attention, which validates the effectiveness of a deeper intra-scale self-attention encoder. A 3-layer multi-scale (denoted by ``3Multi'') self-attention encoder achieves better performance than a 3-layer intra-scale (3Intra) self-attention encoder. It demonstrates that enabling multi-scale information exchange leads to better performance than only conducting intra-scale information exchange alone. However, the large increase of FLOPs by replacing intra-scale with multi-scale self-attention encoder makes us choose a combination of intra-scale and multi-scale self-attention encoders, namely, the design of 2intra-inter-2intra. In the previously mentioned multi-scale encoder, we share both Transformer and FFN weights for features from intra-scale self-attention layers, which reduces the number of parameters and learns common patterns of multi-scale features. It increases the generalization of the proposed SMCA and achieves a better performance of 43.7 with fewer parameters.

\subsection{Overall Performance Comparison}

In Table \ref{tab:final}, we compare our proposed SMCA with other object detection frameworks on COCO 2017 validation set. DETR-R50~\cite{carion2020end} and DETR-DC5-R50 stand for DETR with ResNet-50 and DETR with dilated ResNet-50. Compared with DETR, our SMCA can achieve fast convergence and better performance. Faster RCNN~\cite{ren2016faster} with FPN~\cite{lin2017feature} is a two-stage approach for object detection. Our method can achieve better mAP than Faster RCNN-FPN-R50 at 109 epochs (45.6 vs 42.0 AP). As Faster RCNN uses ROI-Align and feature pyramid with downsampled \{8, 16, 32, 64\} features, Faster RCNN is superior at detecting small objects (26.6 vs 25.9 mAP). Thanks to the multi-scale self-attention mechanism that can propagate information between features at all scales and positions, our SMCA is better for localizing large objects (62.6 vs 53.4 AP). 

Deformable DETR~\cite{zhu2020deformable} replaces the original self-attention of DETR with local deformable attention for both the encoder and the decoder. It achieves faster convergence compared with the original DETR. Exploring local information in Deformable DETR results in fast convergence at the cost of degraded performance for large objects. Compared with DETR, the $\text{AP}_{L}$ of Deformable DETR drops from 61.1 to 58.0. Our SMCA explores a new approach for fast convergence of the DETR by performing spatially modulated co-attention. As SMCA uses global self-attention for information exchange between all scales and positions, our SMCA can achieve better performance for large objects compared with Deformable DETR. Deformable DETR uses downsampled 8, 16, 32, 64 multi-scale features and 8 sampling points for deformable attention. Our SMCA only uses downsampled 16, 32, 64 features and 1 center point for spatial prior. SCMA achieves comparable mAP with Deformable DETR at 50 epochs (43.7 vs. 43.8 AP). As SMCA focuses more on global information and deformable DETR focuses more on local features, SMCA is better at $\text{AP}_{L}$ (60.4 vs 59.0 AP) while inferior at $\text{AP}_{S}$ (24.2 vs 26.4). 

UP-DETR~\cite{dai2020up} can achieve fast convergence and better performance compared with the original DETR due to the exploitation of unsupervised auxiliary tasks. TSP-FCOS and TSP-RCNN~\cite{sun2020rethinking} combines DETR's Hungarian matching with FCOS~\cite{tian2019fcos} and RCNN~\cite{ren2016faster} detectors, which results in faster convergence and better performance than DETR. As TSP-FCOS and TSP-RCNN inherit the structure of FCOS and RCNN that uses local-region features for bounding box detection, they are strong at small objects but weak at large ones. For short training schedules, TSP-RCNN and GMCA-R50 achieve comparable mAP (43.8 at 38 epochs vs 43.7 at 50 epochs), which are better than 43.1 at 38 epochs by TSP-FCOS. For long training schedules, SMCA can achieve better performance than TSP-RCNN (45.6 at 108 epochs vs 45.0 at 96 epochs). TSP use more parameter than SMCA. We observe similar trends by replacing ResNet-50 backbone with ResNet-101 backbone as shown in the lower half part of Table~\ref{tab:final}.

\section{Conclusion}

DETR~\cite{carion2020end} proposed an end-to-end solution for object detection beyond previous two-stage~\cite{ren2016faster} and one-stage approaches~\cite{redmon2016you}. By integrating the Spatially Modulated Co-attention (SMCA) into DETR, the original 500 epochs training schedule can be reduced to 108 epochs and mAP increases from 43.4 to 45.6 under comparable inference cost. SMCA demonstrates the potential power of exploring global information for achieving high-quality object detection. In the future, we will explore SMCA in more scenarios beyond object detection, such as general visual representation learning. We will also explore flexible fusions of local and global features for faster object detection.

\section*{Acknowledgement}
Project supported by the Shanghai Committee of Science and Technology, China (Grant No. 21DZ1100100 and 20DZ1100800)

{\small
\bibliographystyle{ieee_fullname}
\bibliography{egbib}
}

\end{document}